# Generating Benchmark Health Data Using a Tabular Diffusion Transformer

Hao Yan[1,2], Lisa Pilgram[1,2,3], Dan Liu[1,2], Linglong Kong[4], Fida Dankar[1*], Khaled El Emam[1,2.5*]

*[1]Children's Hospital of Eastern Ontario Research Institute, Ontario, Canada*
*[2]School of Epidemiology and Public Health, University of Ottawa, Ontario, Canada*
*[3]Department of Nephrology and Medical Intensive Care, Charité - Universitaetsmedizin Berlin, Berlin, Germany*
*[4]Department of Mathematical and Statistical Sciences, University of Alberta, Edmonton, Canada*
*[5]Woodway Assurance Ltd, Ontario, Canada*

* senior authors

Corresponding Authors:
Khaled El Emam & Fida Dankar
Children's Hospital of Eastern Ontario Research Institute
401 Smyth Road
Ottawa, Ontario K1H 8L1
Canada

{kelemam, fdankar}@ehealthinformation.ca

## Abstract

Cross-Tabular Data Generation (CTDG) seeks to learn a generative model from multiple heterogeneous tables and produce new synthetic tabular datasets. However, existing synthetic tabular data generation methods are largely restricted to single-input-table scenarios and struggle to effectively handle multiple heterogeneous tables with diverse feature sets. To address this limitation, we propose a two-stage framework for cross-tabular data generation. In the first stage, each heterogeneous raw table is transformed into a standardized statistical table with the same set of columns across all tables. Each statistical table captures the marginal distributions of the original columns and the pairwise correlations among them. In the second stage, a diffusion transformer model is trained to capture structural patterns across these homogeneous statistical tables and to generate synthetic statistical tables. Synthetic raw tables are subsequently reconstructed from the generated statistical tables via multivariate Gaussian sampling followed by an inverse probability integral transform. This two-stage CTDG framework enables the learning of a unified generative model from multiple heterogeneous tables and supports the generation of an unlimited number of realistic synthetic heterogeneous tables. Experimental results demonstrate high fidelity in the learned statistical representations and a favorable fidelity–diversity trade-off in the generated synthetic data, validating the effectiveness of the proposed approach.

# 1. Introduction

Interest in and adoption of synthetic tabular data generation (STDG) has been growing quite rapidly over the last few years. SDG is mostly achieved by modeling a joint distribution based on real data and then generating a synthetic dataset from that very model. The resulting synthetic data should mimic the statistical properties and patterns of the real data. This approach enables privacy-preserving data sharing and experimentation, mitigates data scarcity, and supports benchmarking and reproducibility across institutions [1–7]. Modern STDG methods span generative adversarial networks (GANs), variational autoencoders, and more recently, diffusion-based models that often outperform earlier approaches on mixed discrete–continuous datasets [1–3].

In the healthcare domain, synthetic patient records are particularly valuable. They can approximate real cohort distributions, enable cross-site collaborations, simulate rare conditions, and reduce regulatory friction when used appropriately, while still requiring careful evaluation of privacy and utility trade-offs [5–8]. Empirically, tabular GAN variants (e.g., CTGAN, medGAN) and more recent diffusion-based models have achieved strong fidelity and downstream task performance on mixed-type tabular data (e.g., with continuous and categorical features), providing a foundation for developing and testing clinical machine learning pipelines without exposing protected health information [1–3,6–8]. Evaluation studies further demonstrate how synthetic data can be utilized to mitigate data scarcity, reduce bias, and support robustness analyses in both general-purpose and medical domains [4,6,8].

An important use case for STDG is the generation of benchmark datasets. Benchmark datasets serve as a common denominator among researchers and have played a fundamental role in advancing machine learning (ML) research in general [9–12]. There have been various initiatives providing open repositories of datasets that could serve as benchmarks. Most of those, however, are not well suited for benchmarking. One challenge represents the sheer number of datasets available of which usually only a few are used for evaluation. The decisions which ones to use is ultimately taken arbitrarily, so that comparability is rarely possible and selective reporting may be an issue (“cherry-picking”) [13]. Another challenge lies in the provision of raw data which leaves preprocessing to the users (e.g. UCI Machine Learning Repository, Kaggle) [14,15]. This is not only time-consuming but can also vary greatly depending on the user for the same data set. Consequently, reproducibility and comparability are not even given when using the same dataset from the repository. A more standardized example of real-world data benchmarks is the Penn Machine Learning Benchmark (PMLB) [16]. This repository consists of preprocessed datasets in a common format (e.g., the same column name for outcome) to facilitate automated benchmarking. The datasets, however, do not present with high diversity and are not healthcare specific [16].

The Diverse and Generative ML benchmark (DIGEN) accounts for these drawbacks and generates (synthetic) benchmarks to offer reproducible benchmarking [17]. Another approach that uses synthetic benchmarks is presented in [18]. However, these datasets are not based on real data, and therefore there is limited assurance that the marginal distributions, correlational structure, patterns of missingness, and variable cardinality are reflective of actual health datasets. This can affect the meaningfulness of results from studies that use these data.

In the current study we develop and evaluate a model for learning patterns from a large number of tabular datasets simultaneously, and generating benchmark data that are reflective of the plurality of these inputs. Learning from multiple tables enables the development of a unified generative model capable of producing synthetic tables that reflect the diversity of real datasets, rather than overfitting to the idiosyncrasies of a single table. Furthermore, aggregating information across tables allows the model to capture shared statistical characteristics and structural patterns within a domain, such as common correlation structures, distributional shapes, and feature interactions, that may not be fully observable in any individual dataset. This is particularly important when individual tables are small, noisy, or biased, as cross-table learning can improve robustness and reduce variance by pooling complementary information.

Moreover, learning from multiple heterogeneous real data tables enhances generalization: the model can generate new synthetic tables with realistic but previously unseen combinations of features and distributions, allowing benchmarking under varied conditions. It also facilitates knowledge transfer across datasets, enabling the model to incorporate signals from high-quality or data-rich sources into settings where data is scarce. Collectively, these benefits motivate the need for Cross-Tabular Data Generation (CTDG), where the goal is to learn a generative model from multiple heterogeneous tables and produce diverse, realistic synthetic datasets that reflect both shared domain structure and source-specific variation.

# 2. Background

## 2.1 Synthetic tabular data generation

STDG aims to learn generative models from tabular records that capture the statistical relationships among features and enable the creation of realistic synthetic datasets. Existing STDG methods are primarily designed for a single input table, where each record is treated as an independent sample and the model learns a distribution over rows within that fixed schema. As a result, these methods inherently assume a homogeneous feature space and are not equipped to learn from multiple heterogeneous tables whose columns, data types, and semantics differ across datasets. This limitation prevents them from leveraging the broader information available in real-world settings, where data is typically distributed across collections of related but non-aligned tables.

Contemporary STDG methods can be broadly categorized according to their underlying model structures. One type of approaches rely on statistical or classical machine learning models, such as Gaussian Copula [4], Sequential Decision Trees [19], Bayesian Networks [20], and Adversarial Random Forests (ARFs) [21], to approximate the joint distribution of tabular data. Subsequent methods adopt deep generative architectures, including Generative Adversarial Networks (GANs) [22] and Variational Autoencoders (VAEs) [23]. Representative examples include Conditional Tabular GAN (CTGAN) [1], Tabular VAE (TVAE) [1], and Robust Tabular VAE (RTVAE) [24], which enhance generative fidelity and robustness for heterogeneous tabular data.

More recent studies leverage the powerful data modeling capability of diffusion models [25] for tabular generation, such as the Tabular Denoising Diffusion Probabilistic Model (TabDDPM) [2] and TabSyn [3], which achieve state-of-the-art performance in fidelity and privacy protection. In parallel, another research line reformulates tabular data generation as a sequence modeling problem, where each record is treated as a sentence and large-scale language models are fine-tuned for synthetic data generation. Examples include TabMT [26] and GReaT [27], which explore the adaptability of transformer [28] -based language models to structured tabular domains. However, the standard STDG setting focuses on learning from a single table, making it unsuitable for scenarios that involve learning from multiple heterogeneous tables.

## 2.2 Cross table learning

Cross-Table Learning aims to learn from multiple heterogeneous tables for either discriminative or generative purposes. The importance of this problem has been emphasized in a recent study [29], which outlines key requirements for developing tabular foundation models, including support for mixed-type columns, cross-dataset modeling, textual context integration, and permutation invariance.

Recent work [30] has explored cross-table learning through a schema-aware autoencoder and a conditional latent diffusion process, introducing a diffusion-based generative foundation model. While promising, this approach pretrains a generative model on multiple tables but subsequently fine-tunes and evaluates the model on individual tables, with generative performance assessed only in single-table settings rather than in terms of its ability to generate diverse heterogeneous datasets. Moreover, it relies heavily on metadata and column names, which limits its capacity to produce new datasets that follow domain-specific distributions in the absence of schema information.

Most existing works focus on discriminative tasks, such as classification and regression. TaBERT [31] and TURL [32] adapt the BERT [33] architecture for table understanding and parsing, addressing single-table classification and regression. XTab [34] extends this line of research through cross-table pretraining of large transformer models for predictive tasks, while CM2 [35] introduces a cross-table masked pretraining framework via prompt-based masked table modeling, also targeting prediction. On the generative side, CTSyn [30] proposes a foundation model for cross-tabular pretraining and single-table generation; however, it was evaluated primarily on single-table generation. Notably, all previous approaches operate at the record level, where each row (record) is treated as an independent instance. In contrast, our proposed method functions at the table level, where each table is treated as a single instance, enabling the model to learn statistical properties and structural patterns across multiple heterogeneous tables.

# 3. Methods

This section presents the proposed two-stage Cross-Tabular Data Generation (CTDG) method. We denote the $N_r$ real raw tables as $\{X_i\}_{i=1}^{N_r}$, where $X_i$ represents the $i$-th table with $C_i$ columns, i.e., $X_i = \{x_1, x_2, \cdots, x_{C_i}\}$. Each column $x_j$ contains $R_i$ rows, expressed as $x_j = \left[x_{1j}, x_{2j}, \cdots, x_{R_i j}\right]^{\mathrm{T}}$.

In the first stage, each heterogeneous real table is converted into a homogeneous statistical table by summarizing (1) the marginal distribution of each column and (2) the correlation matrix among all columns. Each statistical table therefore encapsulates the statistical characteristics of its corresponding real table, enabling seamless reconstruction of the original data.

In the second stage, a Tabular Diffusion Transformer (TDT) model is trained on the set of statistical tables to learn their joint distribution, allowing it to generate new synthetic statistical tables. Subsequently, synthetic raw tables are reconstructed from these generated statistical tables. The overall framework of the proposed method is illustrated in Figure 1.

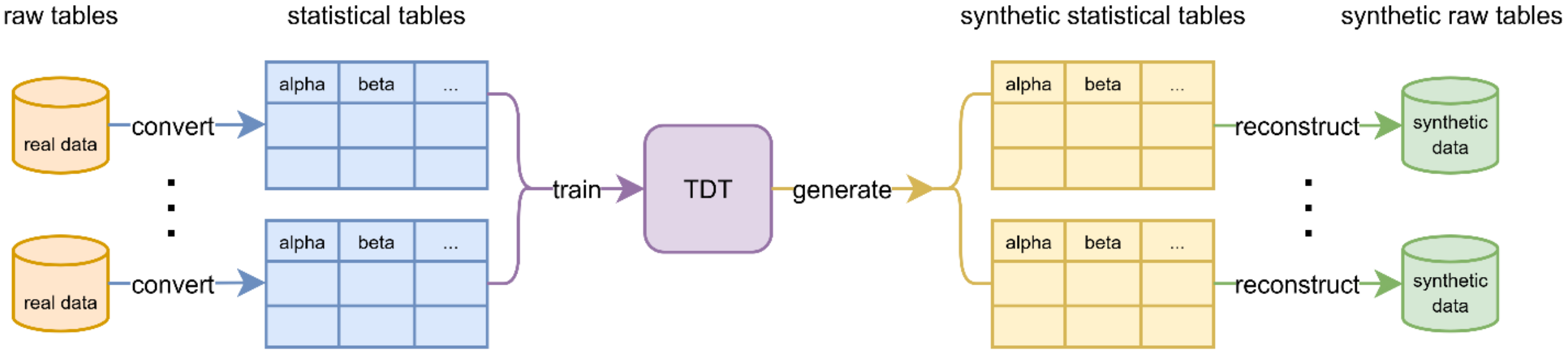

**Figure 1:** The Framework of the proposed method

## 3.1 Statistical tables

Real-world tabular datasets are typically heterogeneous, i.e., they contain varying sets of features across different tables, which makes learning a unified generative model highly challenging. To address this, we converted each real table into a statistical table that summarizes two key aspects: (1) the marginal distribution of each column and (2) the correlations among columns. This transformation allows the model to learn the statistical characteristics shared across multiple tables while disregarding specific column names and actual values—retaining only the approximated table-level statistics necessary for generation.

### 3.1.1 Marginal statistics

Finding a unified distribution for the original columns across multiple tables is non-trivial. First, both intra-table and inter-table columns are heterogeneous, leading to distributions of diverse shapes. Second, directly fitting distributions to numerical features may yield curves with an unknown number of peaks; constraining the number of peaks would, in turn, reduce precision. More importantly, however, preserving the exact categories of categorical features is unnecessary, since the generated synthetic tables are not directly tied to the original columns or category definitions.

To address these issues, we systematically standardize the columns to extract comparable marginal statistics. For each categorical column, we first identify all unique elements and count their occurrence frequencies. The unique elements are then sorted in descending order of frequency and assigned integer indices starting from zero. This unique-to-index mapping is subsequently applied to the entire column, replacing each category with its corresponding index. A Beta–Binomial distribution is then fitted to these mapped indices, resulting in three parameters—$\alpha, \beta$, and the column's cardinality. The Beta–Binomial distribution is chosen because it provides a compact yet flexible parametric representation of categorical frequency patterns, capturing the overdispersion and imbalance commonly observed in real-world categorical data while maintaining a fixed-dimensional summary suitable for downstream generative modeling and reconstruction. An illustrative example of this process is shown in Figure 2.

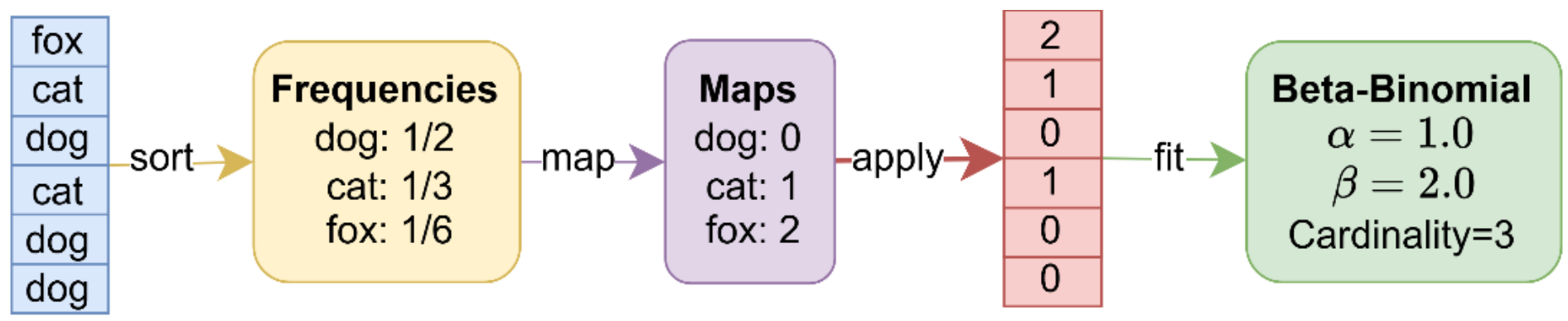

**Figure 2:** Illustration of extracting marginal distribution for categorical features

Numerical features are processed in a similar manner. Their values are first discretized into a fixed number of intervals, $I_N$, after which the discretized intervals undergo the same sorting, mapping, and fitting procedure. This results in parameters $\alpha, \beta$, and a categorical cardinality equal to $I_N$. All subsequent operations are performed on these mapped representations, including the extraction of statistical tables in the first stage and the reconstruction of raw tables from synthetic statistical tables in the final stage. Consequently, we do not aim to reconstruct categorical features using their original category labels or numerical features within their original value ranges; rather, our objective is to learn and reproduce the underlying statistical patterns and distributions shared across multiple tables.

Missing values are also common in tabular data, where certain entries are absent. To account for this, the missing rate of each column is recorded in the corresponding statistical table. Consequently, the parameters describing the marginal distribution of each column, whether categorical or numerical, are represented as ($\alpha, \beta$, cardinality, missing rate). This unified representation provides a compact and consistent summary of column-wise marginal statistics across heterogeneous tables.

### 3.1.2 Representation of correlational structure

Columns within each mapped table are typically interdependent, and preserving these correlations is essential for accurately reconstructing the original mapped table from its corresponding statistical representation. Although pairwise correlations could be computed directly on the mapped columns, sampling from a multivariate Beta–Binomial distribution during reconstruction is computationally challenging. In contrast, sampling from a multivariate Gaussian distribution is both efficient and well supported. We transformed each mapped column to a normal distribution using the probability integral transform [36], which preserves the marginal ordering of the data. Pairwise correlations are then computed among these transformed normal variables, yielding a Gaussian correlation structure that enables efficient joint sampling while maintaining the original dependency relationships.

Since the raw tables contain varying numbers of columns, their corresponding correlation matrices also have different dimensions. Denote the correlation matrix of table $X_i$ as $\Sigma_i$, which has a dimension of $C_i \times C_i$. One might consider using each row of the correlation matrix as a representation of a column's correlation with all other columns. However, this approach would lead to an inconsistent number of correlation parameters across tables with different column counts. Moreover, when generating synthetic statistical tables, such an approach would not ensure that the resulting correlation matrices are symmetric, as the pairwise relationships are not explicitly enforced.

To overcome these issues, we perform eigen-decomposition (i.e., spectral decomposition) of the symmetric correlation matrix. Specifically, each correlation matrix $\Sigma_i$ is decomposed as

$$\Sigma_i = U\Lambda U^\top = U\Lambda^{1/2}\left(U\Lambda^{1/2}\right)^\top = AA^\top,$$

where $U \in R^{C_i \times K}$ contains orthonormal eigenvectors, and $\Lambda \in \mathrm{R}^{K \times K}$ is a diagonal matrix containing the corresponding eigenvalues. The hyperparameter $K$ specifies a fixed upper bound on the dimensionality of the spectral representation and is fixed across all tables to ensure representational consistency. Rather than performing a standard low-rank approximation, we adopt an over-parameterized setting in which $K$ is chosen to be sufficiently large to accommodate tables with a wide range of column counts, thereby minimizing precision loss for tables with many columns.

For tables with fewer than $K$ columns, zero-padding is applied to $U$ to maintain dimensional consistency without introducing spurious correlations. For tables with more than $K$ columns, the column dimension is reduced by randomly dropping columns so that exactly $K$ columns are retained, preventing extremely wide tables from dominating the representation while preserving a fixed-dimensional interface for downstream modeling.

The resulting matrix $A \in R^{C_i \times K}$, defined as $A = U\Lambda^{1/2}$, serves as a compact yet expressive representation of the correlation matrix, from which the original correlations can be recovered as $\Sigma_i = AA^\top$. Each row vector $\boldsymbol{a}_i$ of matrix $A$ corresponds to the correlation parameters associated with a single column in the raw table. Consequently, the correlation between any two columns $i$ and $j$, in either real or synthetic tables, can be computed as the inner product $\langle \boldsymbol{a}_i, \boldsymbol{a}_j \rangle$.

By combining these correlation parameters with the marginal statistics parameters, each column in the raw table is represented by a parameter tuple ($\alpha, \beta$, cardinality, missing rate, $\boldsymbol{a}$), resulting in a total of $K + 4$ parameters per column. Collectively, these rows of parameters form the statistical table representing the original raw table, thereby completing the transformation from heterogeneous to homogeneous representation.

### 3.1.3 Reconstructing raw tables

We now describe the procedure for reconstructing a raw table from its corresponding statistical table, applicable to both real and synthetic cases. The reconstruction begins by recovering the correlation matrix using the correlation parameters $\boldsymbol{a}$ from all columns. Based on this reconstructed correlation matrix, we sample instances from a multivariate Gaussian distribution with zero mean and the recovered correlation structure. Each sampled column is then mapped to its respective Beta–Binomial distribution through the inverse probability integral transform, using the associated parameters $\alpha, \beta$, and cardinality. Finally, missing values are introduced randomly according to the recorded missing rate for each column.

## 3.2 Learning from multiple tables

Our objective is to learn a generative model from the produced statistical tables and to generate synthetic statistical tables that follow a similar distribution as the real ones. The rows across all statistical tables share the same number of parameters, and each parameter carries a specific semantic meaning, which ensures homogeneity. Each statistical table represents a unique combination of statistical rows corresponding to different columns in the original raw table. The goal is to learn the underlying distribution of these combinations across multiple tables and to generate new combinations that exhibit similar statistical properties, from which synthetic raw tables can subsequently be reconstructed.

Given the homogeneity of rows and the varying numbers of rows across tables, transformer [28] -based architectures naturally serve as an effective backbone model. Each row of a statistical table is treated as an embedding, while all rows within one table form a sequence analogous to a sentence in natural language. Recent studies in synthetic data generation [2] have demonstrated that diffusion models [25] excel at maintaining the fidelity of generated data. Motivated by these observations, we adopt a tabular diffusion transformer model for cross tabular data generation that is capable of learning from multiple homogeneous statistical tables.

### 3.2.1 Standardizing columns

Each row (i.e., embedding) in a statistical table contains $K + 4$ parameters with distinct distributions and value ranges. Specifically, the parameters $\alpha$ and $\beta$ are strictly positive and unbounded, with $\alpha \leq \beta$ enforced by the frequency-based sorting procedure used during marginal statistics extraction. The cardinality parameter is lower bounded by 2 but has no fixed upper limit, while the missing rate lies strictly within the interval $[0, 1)$. The remaining $K$ correlation-related parameters correspond to a row vector of the spectral embedding matrix $A$; each entry is real-valued and bounded within $[-1, 1]$.

Embeddings with such diverse value scales hinder the stable training of deep models. To address this, we apply a global quantile transformation [37] to each column across all statistical tables. Quantile transformation is a nonparametric normalization technique that maps a feature's empirical distribution to a target distribution by matching quantiles: values are first mapped to their empirical cumulative probabilities and then transformed via the inverse cumulative distribution function of a chosen target distribution, preserving rank order while standardizing scale. For instance, a global quantile transformation is fitted to the first column ($\alpha$) using all tables, mapping its values to a standard normal distribution through probability integral transform. Consequently, $K + 4$ global quantile transformations are learned and stored. During generation, these transformations are inverted to map the synthetic values back to their original ranges. After the quantile transformation, all values across statistical tables are standardized to comparable scales, making them suitable for transformer-based modeling.

### 3.2.2 Diffusion transformer

**Motivation**. Traditional diffusion models [25] commonly use U-Net [38] architectures, which are not well suited for handling inputs of variable length. Diffusion transformer [39] architectures were later introduced for image generation and include complex conditioning mechanisms for auxiliary information such as labels or text. In our case, only minimal transformer characteristics are required—specifically, the ability to process variable-length sequences and use self-attention to model correlations among embeddings. Therefore, we adopt a simplified diffusion transformer model as the generative backbone for learning from multiple homogeneous tables.

**Model structure**. The tabular diffusion transformer model employs a standard transformer encoder [28] as its core architecture. The trainable components include an input projection layer, a time projection layer, the transformer encoder module, and an output projection layer. The input projection layer first maps each input sequence to the standard embedding dimension (typically 768). A sinusoidal time embedding function [25] encodes the diffusion timestep, which is projected to the same embedding dimension via the time projection layer and added to the input embeddings. The resulting embeddings from all rows are passed through the transformer encoder to produce output embeddings, which are finally mapped back to $K + 4$ dimensions through the output projection layer. The model structure is illustrated in Figure 3.

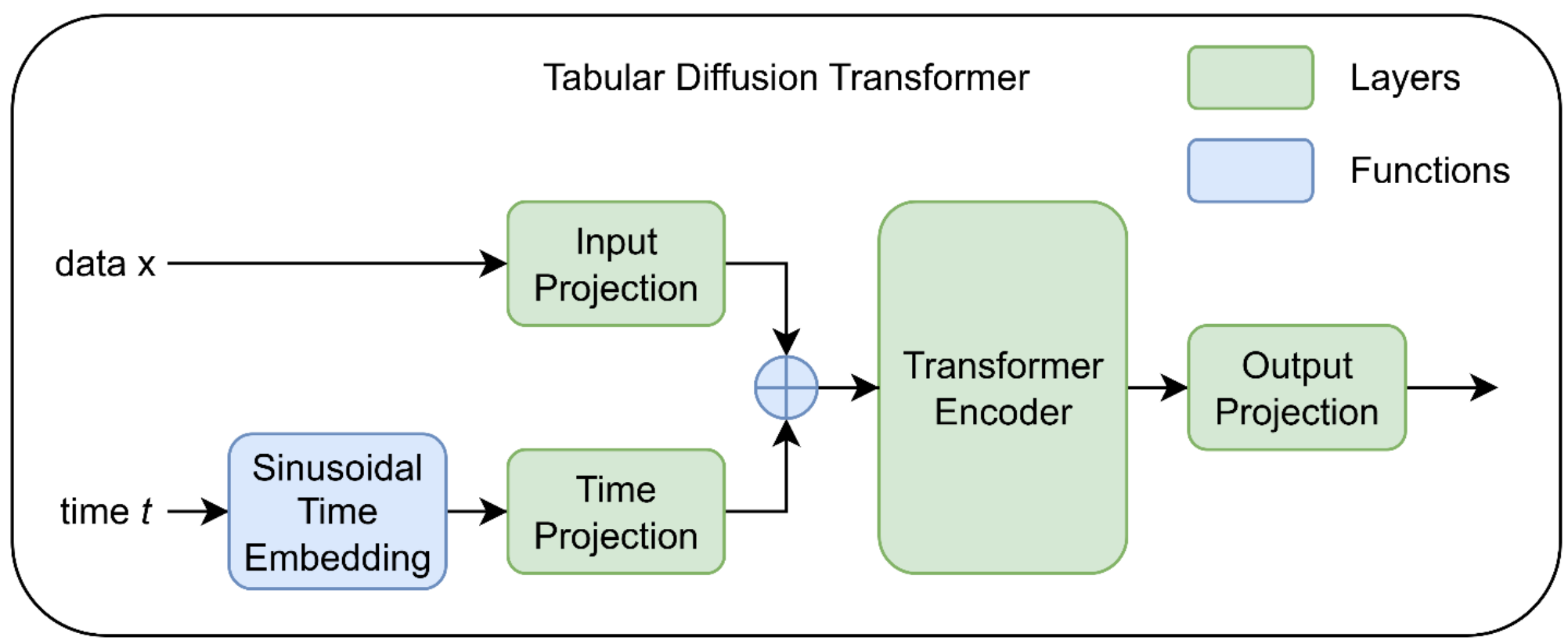

**Figure 3:** Illustration of the tabular diffusion transformer model structure.

**Training**. We follow the standard Denoising Diffusion Probabilistic Model (DDPM) [25,40] training procedure, with the simplified loss function:

$$L_{\text{simple}} = E_{t,\mathrm{x}_0,\boldsymbol{\epsilon}} \|\boldsymbol{\epsilon} - \boldsymbol{\epsilon}_\theta(\mathrm{x}_t, t)\|^2.$$

A statistical table $\mathrm{x}_0$ is first sampled from the dataset, along with a timestep $t \sim \mathcal{U}(1, T)$, and standard normal noise $\boldsymbol{\epsilon} \sim \mathcal{N}(\mathbf{0}, \mathbf{I})$ with the same dimension as $\mathrm{x}_0$. Here, $T$ is a hyperparameter defining the maximum timestep. The noisy sample at timestep $t$ is computed as

$$\mathrm{x}_t = \sqrt{\bar{\alpha}_t}\mathrm{x}_0 + \sqrt{1 - \bar{\alpha}_t}\boldsymbol{\epsilon},$$

where $\bar{\alpha}_t$ is a coefficient derived from a linear diffusion scheduler [25]. The noisy input $\mathrm{x}_t$ and the corresponding timestep $t$ are fed into the diffusion transformer model $\boldsymbol{\epsilon}_\theta$, which aims to reconstruct the original noise $\boldsymbol{\epsilon}$. The model parameters are updated as

$$\theta \leftarrow \theta - \eta \nabla_\theta \|\boldsymbol{\epsilon} - \boldsymbol{\epsilon}_\theta(\mathrm{x}_t, t)\|^2,$$

where $\eta$ denotes the learning rate. The procedure is repeated iteratively until convergence, with batch-wise optimization used for computational efficiency. The algorithm illustration has been shown in DDPM [25].

**Sampling**. After training, the diffusion transformer can generate synthetic statistical tables that follow the learned distribution of real statistical tables through a $T$-step denoising process [25]. The procedure begins with noise $\mathrm{x}_T \sim \mathcal{N}(\mathbf{0}, \mathbf{I})$. For each timestep $t = T, T-1, \cdots, 1$, another noise sample $\mathbf{z} \sim \mathcal{N}(\mathbf{0}, \mathbf{I})$ is drawn, and one-step denoising is computed as

$$\mathrm{x}_{t-1} = \frac{1}{\sqrt{\alpha_t}}\left(\mathrm{x}_t - \frac{1-\alpha_t}{\sqrt{1-\bar{\alpha}_t}}\boldsymbol{\epsilon}_\theta(\mathrm{x}_t, t)\right) + \frac{1-\bar{\alpha}_{t-1}}{\sqrt{1-\bar{\alpha}_t}}\beta_t\mathbf{z},$$

where $\alpha_t$ and $\beta_t$ are scheduler coefficients. At the final step $(t = 1)$, $\mathbf{z} = \mathbf{0}$ to avoid adding additional noise. This iterative denoising process gradually transforms random noise into coherent synthetic samples, with $\mathbf{z}$ introducing variability that enhances the diversity of generated outputs. The algorithm illustration has been shown in DDPM [25].

### 3.2.3 Pretraining and finetuning

Each statistical table serves as an individual training instance, which makes it difficult to train a diffusion transformer from scratch using only a limited number of tables. To mitigate this, we employ a two-stage training strategy. First, the model is pretrained on a large collection of general tabular datasets from OpenML [41], enabling it to learn diverse statistical patterns. Subsequently, the pretrained model is fine-tuned on health datasets to adapt to its specific statistical characteristics. This pretraining–finetuning approach enhances convergence stability and improves generation quality on domain-specific datasets.

## 3.3 Model training

### 3.3.1 Datasets

We evaluate the proposed method using the OpenML datasets [41] and 13 healthcare related datasets, where the details of the 13 datasets are included in the appendix. In particular, the OpenML datasets labeled with the Health tag, together with the 13 healthcare datasets, are treated as representative benchmarks of the health domain. The objective is to learn from these datasets and generate synthetic tables that preserve the statistical distributions observed in the corresponding real data. For pretraining, we select 4,095 verified OpenML datasets that meet the size requirements for the diffusion transformer model. For fine-tuning, we use 131 OpenML datasets with the Health tag in combination with the 13 healthcare datasets. The OpenML collection spans a wide range of domains, including health, economics, engineering, finance, games and statistics, and offers convenient programmatic access via its public interface.

### 3.3.2 Implementation details

For statistical table extraction, numerical features are discretized into a fixed number of intervals $I_N = 100$, regardless of their value range or scale, to ensure a consistent global rule across all tables. This choice reflects a trade-off between representational precision and computational complexity: using 100 intervals is sufficient to capture the overall distributional shape of numerical features while avoiding excessive cardinality that would increase model complexity and memory usage. Moreover, this value is consistent with the typical range of cardinalities observed in categorical features, enabling a more uniform treatment of numerical and categorical columns within the statistical representation.

The hidden dimension used to represent each correlation matrix is set to $K = 256$. As discussed earlier, this value is chosen to accommodate tables with a wide range of column counts while maintaining a fixed-dimensional representation required for cross-table modeling. Setting $K = 256$ provides sufficient capacity to represent correlation structures of moderately wide tables without significant truncation, while remaining computationally tractable. Tables exceeding this dimensionality are truncated to $K$ columns, whereas tables with fewer columns are zero-padded in their representation matrix $A$ to ensure dimensional consistency.

For synthetic statistical table generation, we employ the proposed diffusion transformer model as the generative backbone. The model uses a standard transformer encoder architecture with an input dimension of $K + 4 = 260$ and an embedding dimension of 768. The transformer encoder comprises 12 layers, each containing 12 attention heads. The total diffusion timesteps are set to $T = 1000$, and each timestep is encoded using a standard sinusoidal time embedding. Coefficients $\alpha_t$ and $\beta_t$ are computed using a linear diffusion scheduler. Training is performed using the AdamW optimizer [42] with a learning rate of $1e - 4$, a batch size of 64, and 500 epochs. The same configuration is applied during fine-tuning on the target datasets.

## 3.4 Model Evaluation

Consistent with the two stages of our framework, our evaluation consists of two main parts:

1. Evaluation of the reconstruction of the original data tables from their statistical table representation. The reconstructed raw data tables should have high fidelity to the original raw data tables. This indicates how well the statistical tables are as a representation of the original raw data.

2. Evaluation of the generated individual level health datasets. For this we assess the extent to which the generated health data tables are closer to real health data compared to real data from other domains, such as engineering, economics, and gaming. Furthermore, we evaluate the diversity of the generated health data tables to determine how well they cover the domain of real health data.

For each of these two evaluation we defined a relevant metric as described below.

### 3.4.1 Evaluation metrics

To quantitatively assess fidelity and answer the first question, we employ the dimension-Normalized Wasserstein Distance (dNWD). Each table is treated as a multivariate random variable, where each row represents an instance sampled from its corresponding multivariate distribution. For a given synthetic table, fidelity is evaluated by measuring the distance between the synthetic and real multivariate distributions.

To enable a closed-form computation of the Wasserstein distance, we map the multivariate distributions to multivariate normal distributions via the cumulative distribution function (CDF) using probability integral transform. The distance between two transformed distributions p and q can then be computed as

$$\mathrm{dNWD} = \overline{W_2}(p,q) = \frac{1}{\sqrt{d}} W_2(p,q) = \sqrt{\frac{1}{d}\left\|\mu_p - \mu_q\right\|^2 + \frac{1}{d}\mathrm{Tr}\left(\Sigma_p + \Sigma_q - 2\left(\Sigma_q^{\frac{1}{2}}\Sigma_p\Sigma_q^{\frac{1}{2}}\right)^{\frac{1}{2}}\right)},$$

where $\mu_p$ and $\mu_q$ are the mean vectors, and $\Sigma_p$ and $\Sigma_q$ are the covariance matrices of $p$ and $q$, respectively. Here, $\mathrm{Tr}$ represents the matrix trace, and $\Sigma^{1/2}$ denotes the symmetric square root of the positive semidefinite covariance matrix $\Sigma$. Unlike the standard 2-Wasserstein distance, the proposed normalized version divides both the mean and covariance terms by the dimension $d$, enabling fair comparisons across tables with different numbers of columns.

To answer the second set of questions, for evaluating synthetic statistical table generation by the diffusion transformer model, there is no one-to-one correspondence between real and synthetic statistical tables. Therefore, we adopt the Closest Wasserstein Distance (CWD), defined as

$$\mathrm{CWD} = \frac{1}{N_s}\sum_{j} \min_{i} \overline{W_2}\left(Z_r^i, Z_s^j\right),$$

where $Z_r$ and $Z_s$ denote the sets of real and synthetic statistical tables, respectively, $i$ indexes the real tables, j indexes the synthetic ones, and $N_s$ is the total number of synthetic tables. For each synthetic statistical table, we compute its distance to all real statistical tables and record the minimum value as its closest match. The average of these minimum distances across all synthetic tables is reported as the overall fidelity measure for the diffusion transformer model.

### 3.5 Ethics statement

This project was approved by the CHEO Research Institute Research Ethics Board, protocol number: CHEOREB# 26/100X. All research reported in this study was performed in accordance with relevant guidelines/regulations. This study is a secondary analysis of public data and data that has been anonymized. The requirement for informed consent was waived by the CHEO Research Institute Research Ethics Board.

# 4. Results

## 4.1 Raw table reconstruction

### 4.1.1 Reconstruction from statistical tables

The first stage of the proposed methodology converts real raw tables into statistical tables. To evaluate the fidelity of this conversion, we reconstruct synthetic raw tables from the extracted statistical tables and compare them with their original counterparts. Specifically, we use the Health (including Ours-13 and OpenML Health) datasets as the source of real raw tables, extract their corresponding statistical tables, and then reconstruct synthetic raw tables from these statistical representations. The dimension-Normalized Wasserstein Distance (dNWD) is computed between each pair of real and reconstructed synthetic raw tables, as illustrated in Figure 4.

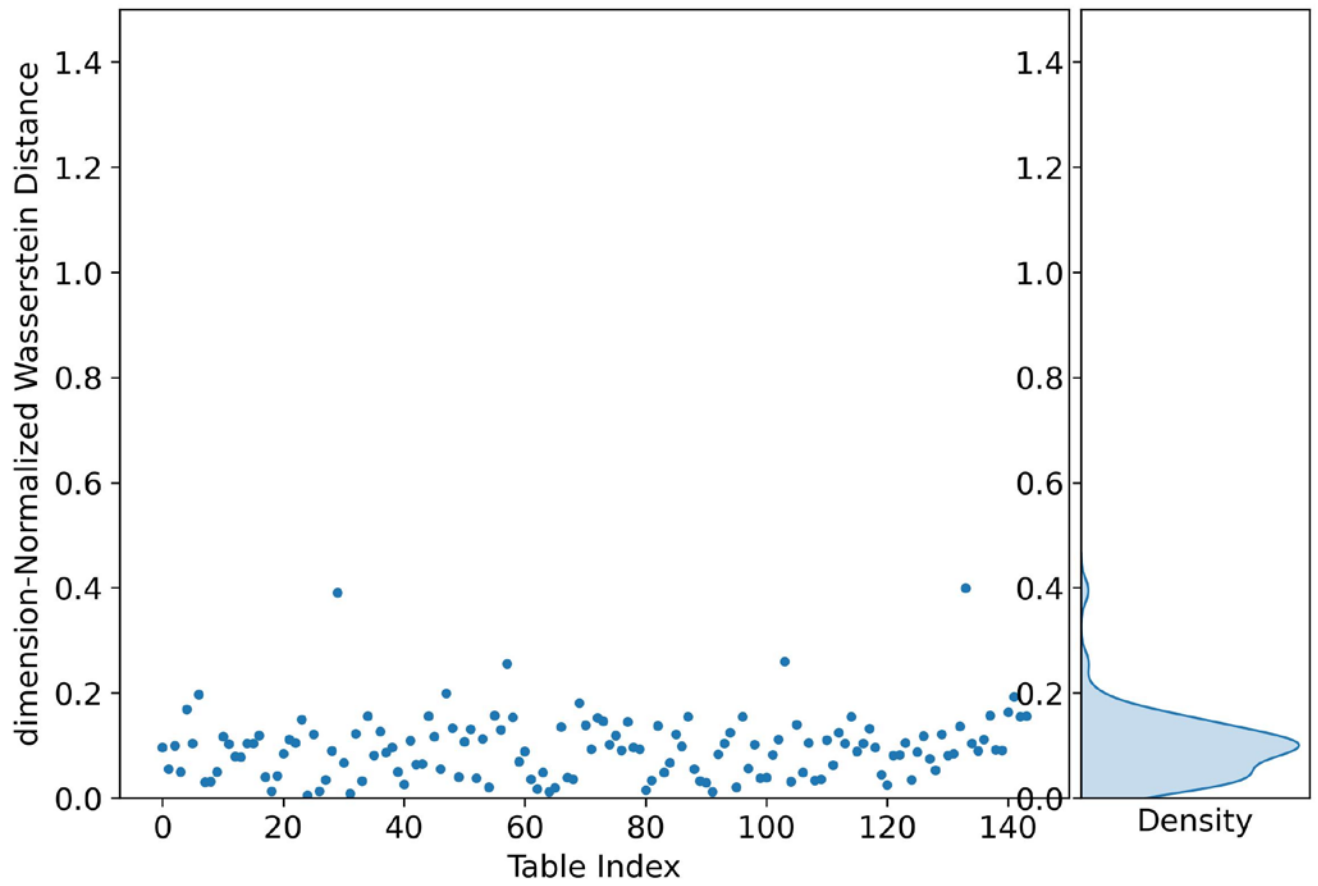

**Figure 4:** The dimension Normalized Wasserstein Distance between each pair of real and synthetic raw tables. The dNWD's across the 144 datasets show an average of 0.095 and standard derivation of 0.061.

The average d-NWD across all datasets is 0.095 with a standard derivation of 0.061. Among the 144 datasets, most achieve d-NWD values below 0.2, while very few fall within the range of 0.2 and 0.4. The non-zero distances mainly come from the fitting of Beta-Binomial distribution where approximation happens due to the smoothness restriction of Beta-Binomial curves. These distances are notably small, suggesting that the reconstructed tables closely match the real tables in terms of their first- and second-order statistics. The results confirm that the reconstructed tables are statistically similar to the originals, demonstrating that the extracted statistical tables serve as reasonably faithful representations of the raw tables.

### 4.1.2 Ablation study: Effect of correlation representation

To evaluate the role of correlation modeling in the statistical representation, we conduct two ablation studies in which the correlation structure is either removed or replaced. In the first setting, we use an identity correlation matrix, which preserves only marginal statistics and assumes independence among features. In the second setting, we use a random correlation matrix, where diagonal elements are fixed to one and off-diagonal elements are independently sampled from $\mathcal{N}(0,1)$, symmetrized, and clipped to the range $[-1,1]$. In both cases, synthetic raw tables are reconstructed from these modified statistical representations and compared with the original raw tables using the dimension-normalized Wasserstein Distance (dNWD).

The results for the identity correlation setting are shown in Figure 5. The average dNWD increases to 0.304 with a standard deviation of 0.164, compared to 0.095 ± 0.061 achieved by the full model. The distances are noticeably larger and more dispersed, indicating that preserving only marginal statistics is insufficient for accurately reconstructing the original tables. This degradation arises from the loss of dependency information between features, which plays a critical role in defining the joint distribution.

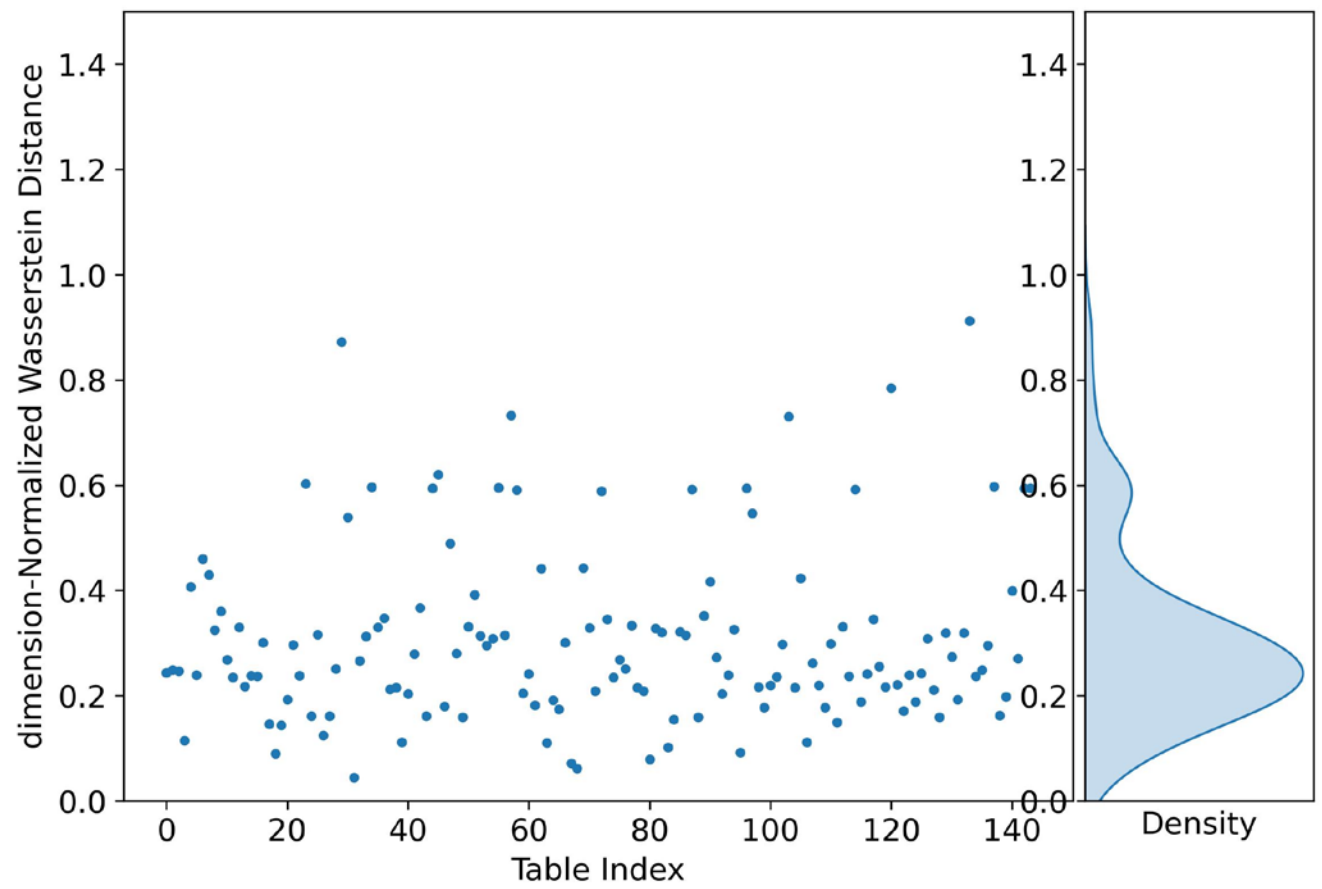

**Figure 5:** Dimension-normalized Wasserstein Distance (dNWD) between real and reconstructed raw tables using only marginal statistics (identity correlation matrix).

The results for the random correlation setting are shown in Figure 6. The reconstruction quality further deteriorates, with an average dNWD of 0.636 and a standard deviation of 0.270. The distribution of distances becomes significantly broader, with many datasets exhibiting large reconstruction errors. This indicates that introducing arbitrary correlations not only fails to recover the true structure, but can actively distort the data by imposing incorrect dependencies between features.

Comparing the two ablation settings reveals a clear pattern: the identity case underestimates dependencies, while the random case introduces spurious ones. Both lead to substantial degradation in reconstruction quality, with the random correlation setting performing the worst. These results demonstrate that accurate modeling of correlations is essential. It is not sufficient to include correlations in the representation; they must reflect the true dependency structure of the data in order to preserve the statistical properties of the original tables.

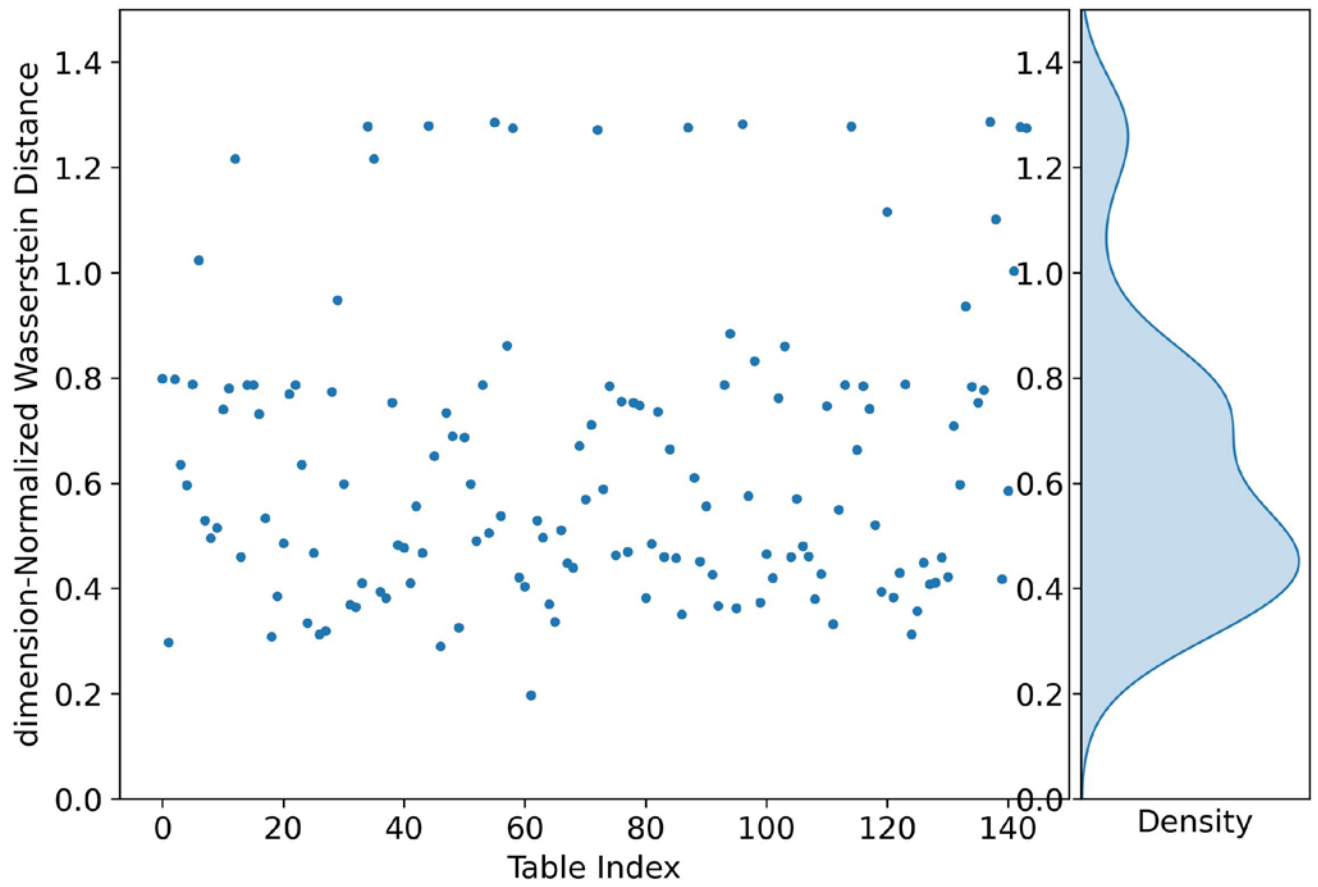

**Figure 6:** Dimension-normalized Wasserstein Distance (dNWD) between real and reconstructed raw tables using a random correlation matrix.

## 4.2 Synthetic statistical table generation

In the second stage of the proposed methodology, we train a diffusion transformer model to learn the distributional and structural patterns of statistical tables drawn from multiple real datasets. The model is first pretrained on a large and diverse collection of real statistical tables and subsequently fine-tuned using statistical tables derived from datasets in the Health domain. After fine-tuning, the model is used to generate new synthetic statistical tables intended to reflect the statistical characteristics of Health-related data. To further assess whether the model captures domain-specific structure rather than generic patterns, we also compare the generated synthetic Health tables against real tables from other OpenML domains.

### 4.2.1 Synthetic-to-real evaluation

To evaluate generalization performance, we conducted 5-fold cross-validation for the tabular diffusion transformer model using 144 datasets from the Health domain. In each fold, 80% of the datasets were used for tuning and the remaining 20% for testing. To ensure a fair comparison across domains, any overlapping tables between domains were removed prior to evaluation.

The quality of the generated synthetic tables was evaluated using the Closest Wasserstein Distance (CWD) between synthetic and real statistical tables. The CWD involves minimization over $N$ distances and is sensitive to the value of $N$. Therefore, for comparability across domains, we fixed the number of real datasets in each domain to 29, matching the size of the Health test set in each fold. In addition to measuring distances between synthetic and real Health tables, we also compute distances between synthetic Health tables and real tables from other domains to verify that the generated data remains closer to its target domain. The results are presented in Table 1. Each row corresponds to a group of real datasets, and each column represents one cross-validation fold.

Across all five folds, the synthetic tables consistently achieve the smallest Closest Wasserstein Distance when compared to real Health tables. In each fold, the average CWD for Health is lower than that for Economics, Statistics, Engineering, Finance, and Games. The five-fold average further highlights this separation: Health attains 0.72, while the next closest domains, Economics and Statistics, yield averages of 0.97 and 0.98, respectively. The remaining domains exhibit substantially larger distances, exceeding 1.3 on average.

**Table 1:** Closest Wasserstein Distance between 2000 synthetic tables and 29 (or 28) real tables under 5-fold cross validation.

| Domain | Fold 1 | Fold 2 | Fold 3 | Fold 4 | Fold 5 | Avg (5-fold) |
|---|---|---|---|---|---|---|
| **Health** | 0.73±0.32 | 0.72±0.34 | 0.71±0.36 | 0.74±0.45 | 0.68±0.33 | **0.72** |
| **Economics** | 0.97±0.32 | 0.96±0.31 | 0.92±0.30 | 1.02±0.34 | 0.96±0.32 | 0.97 |
| **Statistics** | 1.01±0.39 | 1.01±0.39 | 1.03±0.42 | 0.93±0.35 | 0.94±0.36 | 0.98 |
| **Engineering** | 1.42±0.82 | 1.43±0.83 | 1.57±0.88 | 1.12±0.65 | 1.22±0.71 | 1.35 |
| **Finance** | 1.65±1.13 | 1.66±1.15 | 1.80±1.23 | 1.30±0.90 | 1.34±0.99 | 1.55 |
| **Games** | 1.84±1.12 | 1.86±1.13 | 1.98±1.22 | 1.52±0.90 | 1.56±0.99 | 1.75 |

The relative ordering between domains is stable across folds. Economics and Statistics consistently form a second tier with moderately larger distances than Health, whereas Engineering, Finance, and Games display progressively greater discrepancies. This consistent ranking across different train–test splits indicates that the observed proximity between synthetic tables and real Health data is systematic rather than dependent on a particular fold.

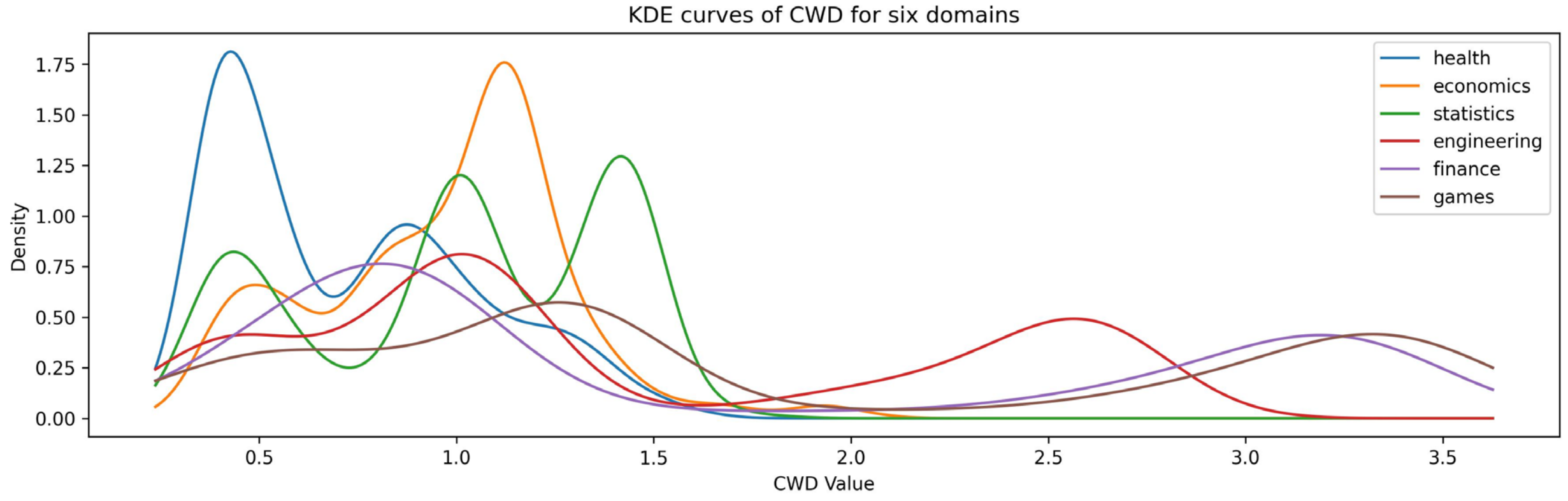

**Figure 7:** Kernel Density Estimation (KDE) curves of CWD values for synthetic-to-real evaluation (Fold 1).

Figure 7 further illustrates this separation at the distributional level. The kernel density estimation (KDE) curves show the empirical distribution of CWD values between the 2000 synthetic tables and each domain for a representative fold. The distribution corresponding to the Health domain is clearly left-shifted, with most probability mass concentrated at smaller CWD values. In contrast, Economics and Statistics exhibit moderately right-shifted distributions, while Engineering, Finance, and Games display progressively heavier right tails and substantially larger support. In particular, Finance and Games extend to much larger CWD values, indicating greater structural divergence from synthetic Health tables. The ordering observed in Table 3 is therefore preserved across the full distribution, demonstrating that the smaller average CWD for Health reflects a systematic shift rather than being driven by a small number of favorable cases.

At the same time, the KDE curves span a relatively wide range of CWD values across all domains, including Health. This spread indicates that the generated synthetic tables are diverse rather than collapsing to a narrow region of the Health data space. While the bulk of synthetic tables align closely with real Health tables, the presence of broader support suggests variability in the generated statistical structures. Such diversity is desirable, as it reflects the model’s ability to capture heterogeneous patterns within the Health domain rather than reproducing only a limited subset of statistical configurations.

### 4.2.2 Hypothesis testing

To formally evaluate whether synthetic Health tables are significantly closer to the real Health domain than to the other domains, we performed paired one-sided Wilcoxon signed-rank tests within each fold.

For fold $f$ and comparison domain $d \neq \mathrm{Health}$, paired differences were computed across 2000 synthetic tables $s$:

$$\Delta_{f,d,s} = \mathrm{CWD}_d(f,s) - \mathrm{CWD}_{\mathrm{Health}}(f,s).$$

The hypotheses were defined as:

$$H_0 : \mathrm{median}\left(\Delta_{f,d,\cdot}\right) = 0,$$
$$H_1 : \mathrm{median}\left(\Delta_{f,d,\cdot}\right) > 0,$$

which test whether synthetic table $s$ is, on average, closer to Health domain than to domain $d$.

This procedure yields one p-value $p_{f,d}$ per fold for each domain $d$. Because cross-validation folds are not fully independent as training sets partially overlap across folds, we do not treat fold-level results as independent replicates, nor do we pool all synthetic tables across folds into a single test. Instead, we regard each fold as a repeated evaluation under a different train–test split and summarize the evidence across folds.

To obtain an overall measure of evidence while respecting the cross-validation structure, we combined the fold-level p-values using Fisher's method:

$$X_d = -2\sum_{f=1}^{5} \ln\left(p_{f,d}\right),$$

which provides a global test statistic for each domain $d$. Although Fisher's method formally assumes independence under the null hypothesis, in this setting it serves as a conservative aggregation of consistent fold-level evidence rather than as an exact variance-based estimator. More importantly, the directional effect is consistent across folds: the median paired differences $\Delta_{f,d,\cdot}$ are positive in all five folds for every domain $d$, indicating uniform support for the alternative hypothesis across cross-validation splits.

To account for multiple comparisons across the five non-Health domains, we applied the Holm step-down procedure at a family-wise significance level of $\alpha = 0.05$. For all domains, the combined Fisher statistics were extremely large, corresponding to $\log_{10}$ p-values below −500 in every case, and all comparisons remained significant after Holm correction. Taken together, the consistent fold-level effects and the aggregated statistical evidence demonstrate that the synthetic tables generated after Health-domain fine-tuning are significantly closer to real Health tables than to tables from any other evaluated domain under the Closest Wasserstein metric.

### 4.2.3 Real-to-real domain distance analysis

To provide a real-data reference for the proximity structure observed in the synthetic evaluation, we computed the Closest Wasserstein Distance (CWD) between the real Health training data and real datasets from each domain under the same 5-fold cross-validation splits. For comparability, the number of datasets per domain was fixed to 29, matching the size of the Health test set. The results are shown in Table 2.

**Table 2:** Closest Wasserstein Distance between real Health training set and real datasets from each domain.

| Domain | Fold 1 | Fold 2 | Fold 3 | Fold 4 | Fold 5 | Avg (5-fold) |
|---|---|---|---|---|---|---|
| **Health** | 0.38±0.23 | 0.44±0.32 | 0.41±0.32 | 0.46±0.48 | 0.43±0.33 | **0.42** |
| **Economics** | 0.60±0.40 | 0.61±0.43 | 0.60±0.42 | 0.64±0.44 | 0.62±0.38 | 0.61 |
| **Statistics** | 0.61±0.45 | 0.62±0.45 | 0.60±0.44 | 0.64±0.46 | 0.62±0.44 | 0.62 |
| **Engineering** | 0.76±0.77 | 0.81±0.84 | 0.80±0.83 | 0.83±0.83 | 0.81±0.77 | 0.80 |
| **Finance** | 0.89±0.97 | 0.95±1.07 | 0.96±1.05 | 0.99±1.06 | 0.91±0.98 | 0.94 |
| **Games** | 0.98±1.03 | 1.05±1.12 | 1.02±1.11 | 1.07±1.12 | 1.01±1.03 | 1.03 |

The same ordering observed in the synthetic-to-real evaluation appears here. Health–Health distances are smallest (five-fold average 0.42), followed by Economics and Statistics, with Engineering, Finance, and Games progressively more distant. The ranking is consistent across folds, indicating a stable inter-domain similarity hierarchy in the real data itself.

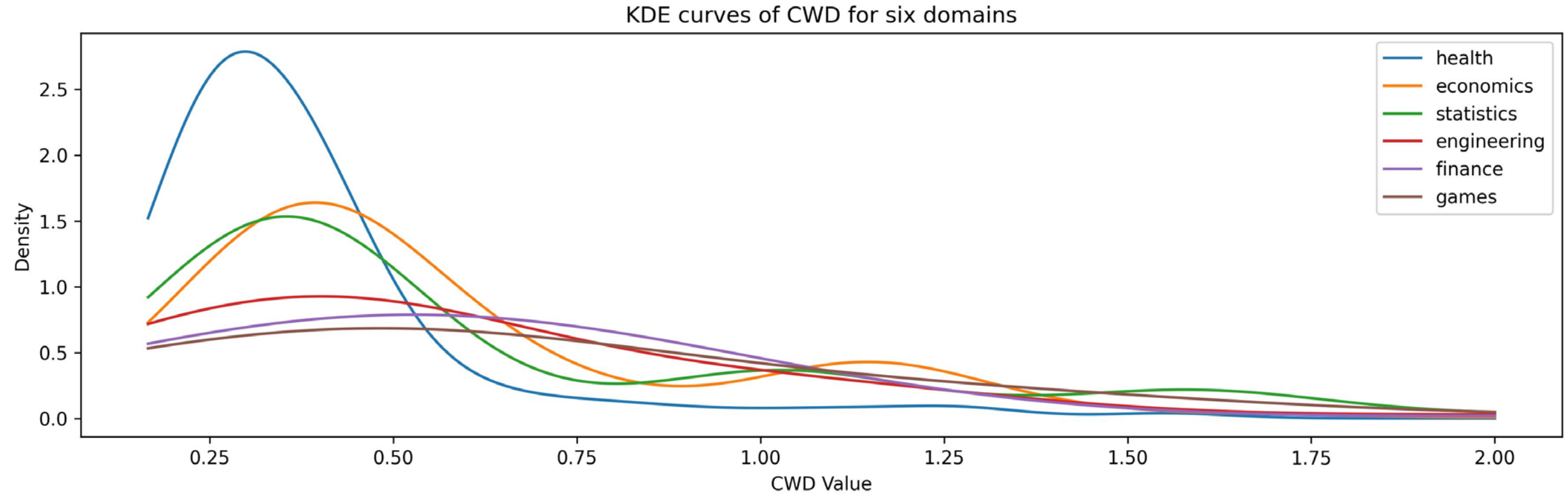

**Figure 8:** Kernel Density Estimation (KDE) curves of the CWD values for real-to-real evaluation (Fold 1).

Figure 8 visualizes the distribution of these distances for a representative fold. The Health–Health distances are concentrated at smaller values and are clearly left-shifted relative to cross-domain distances. As domains become less related to Health, the distributions shift further to the right and become more dispersed. This confirms that the separation observed in Table 2 reflects a systematic distributional shift rather than isolated outliers.

Statistical testing further supports this structure. Within each fold, paired one-sided Wilcoxon signed-rank tests show that Health–Health distances are significantly smaller than Health–other distances. Combining fold-level p-values using Fisher's method and applying Holm correction across domains yields significant results for all comparisons (combined p-values between $3.6 \times 10^{-18}$ and $3.6 \times 10^{-62}$).

More importantly, this intrinsic similarity structure in real data mirrors the pattern observed in the synthetic-to-real evaluation. In both settings, Health is closest to Health, Economics and Statistics form a second tier, and Engineering, Finance, and Games are progressively further away. The synthetic model therefore preserves not only absolute closeness to the Health domain, but also the relative cross-domain proximity relationships present in real datasets. This alignment suggests that fine-tuning on Health data enables the diffusion model to capture domain-specific structural characteristics while maintaining the broader similarity geometry among domains.

#### 4.2.4 Diversity analysis

To further assess the diversity of the generated synthetic statistical tables, we analyze how synthetic samples are distributed with respect to the real Health datasets. For each synthetic table, the Closest

Wasserstein Distance (CWD) is computed against all real tables in the Health domain, and we record the indices of the top-$k$ nearest real tables.

If the synthetic tables lack diversity (e.g., mode collapse), many synthetic samples would map to only a small subset of real tables, even when considering multiple nearest neighbors. Conversely, if the synthetic tables are diverse and well distributed, the set of Top-$k$ nearest neighbors across all synthetic tables should cover a large proportion of the available real tables. Based on this intuition, we measure diversity using the Top-$k$ Ratio, defined as the proportion of unique real table indices that appear among the top-$k$ nearest neighbors across all synthetic tables. The results aggregated across all folds are shown in Figure 9, where we report the Top-$k$ Ratio for $k = 1, \ldots, 10$.

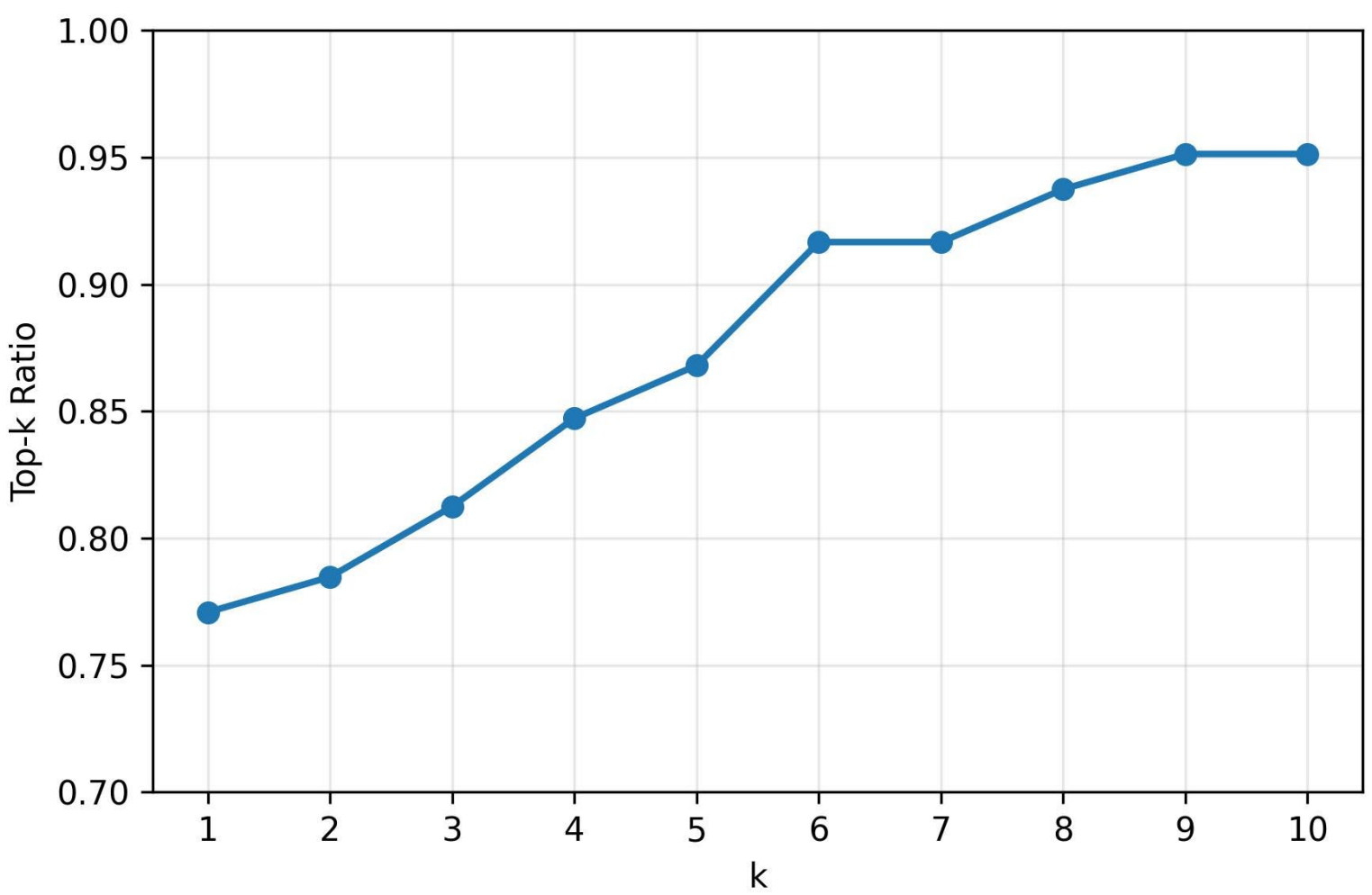

**Figure 9:** Top-$k$ diversity of synthetic tables measured by the proportion of unique real Health tables covered by the top-$k$ nearest neighbors (in terms of WD) across all synthetic samples.

For $k = 1$, the ratio is 0.77, indicating that the nearest neighbors of synthetic tables already cover 77% of the real tables. As $k$ increases, the coverage improves steadily, reaching 0.87 at $k = 5$ and exceeding 0.95 at $k = 9$ and $k = 10$. This trend shows that synthetic tables are not concentrated around a small subset of real tables, but instead are broadly distributed across the domain.

The increase from Top-1 to higher $k$ also reveals additional structure: while each synthetic table has a closest match, many have multiple nearby real counterparts, suggesting that the generated tables lie in regions that are well supported by the real data distribution rather than being isolated or degenerate.

The saturation of the ratio near 0.95 further indicates that almost all real tables are represented within the neighborhood structure of the synthetic samples.

These findings are consistent with the KDE analysis in Figure 5, where the CWD distributions span a wide range of values. Together, they demonstrate that the model avoids mode collapse and captures diverse statistical patterns within the Health domain. Combined with the strong alignment observed in the synthetic-to-real evaluation, this suggests that the model achieves both fidelity (closeness to real Health data) and diversity (broad coverage of the domain).

# 5. Discussion

## 5.1 Summary

The objective of the current paper was to generate large scale realistic health datasets that can be used for benchmarking studies. We address this by developing a model for cross tabular data generation, in which a generative model is learned from multiple heterogeneous tabular datasets and used to generate new, plausible synthetic heterogeneous tables. We propose a two-stage methodology. In the first stage, heterogeneous raw tables are transformed into homogeneous statistical tables that capture both marginal and correlation statistics. In the second stage, a diffusion transformer model is trained on these homogeneous statistical tables to learn their underlying distribution and generate new synthetic statistical tables. Synthetic raw tables can then be reconstructed by sampling from the generated statistical tables. Experimental results demonstrate that the proposed approach achieves high fidelity in reconstructing raw tables and produces synthetic statistical tables that exhibit both strong fidelity to real health data and meaningful diversity. These provide evidence supporting effectiveness of the proposed methodology in learning from multiple heterogeneous tables and generating new, statistically consistent synthetic tabular data.

An important consideration when generating synthetic data is privacy, and the extent to which the generated data pose privacy risks to the data subjects in the original datasets used for training [43]. The representation of a dataset is the statistical table, which is an aggregate description of the distributions and correlational structure of the individual-level datasets. It is this representation that is used to train the generative model. Even if the trained and tuned diffusion model faithfully reconstructs the statistical table, there will not be a one-to-one mapping between records in the training data and individual data produced from the statistical table, and meaningful memorization could only occur in degenerate cases where feature values are constant, in which case sampling from the marginal distribution (a beta-binomial distribution) would produce the same original values. Furthermore, that there is not a one-to-one mapping nor concordance between the features themselves in the training datasets and the datasets generated precludes assessments of membership and attribute disclosure.

Future work should further explore the representation of raw data tables. In our work we used statistical tables, however, other embedding techniques may be evaluated as they may retain more information about the structure of the datasets.

## 5.2 Applications in practice

We used the generative model to provide 1000 datasets on OSF[1]. Each dataset has 10,000 rows and the number of columns varies between 10 and 256. These can be used to perform benchmark and simulation studies that require realistic health datasets. For example, the 1000 datasets can be used to compare the performance of multiple generative models for tabular data at scale (where previous studies only considered at most a handful of health datasets). By adding labels to these datasets reflecting different outcome balance, the dataset can be used to simulate the performance of machine learning classifiers on plausible health datasets.

Our generated datasets are expected to produce more realistic benchmarking results than previous benchmark datasets for health-related situations in that the generation is based on real health data to start off with rather than hypothetical distributions. The latter is the approach used in previous studies [17], [18].

The methodology that we have presented in this study can be applied in other domains for the generation of benchmark datasets. This would make plausible domain-specific datasets for future studies readily available at scale.

## 5.3 Limitations

The statistical table representation may mask the effect of outliers in the original data. While the modeling of marginal distributions can capture those with heavy tails, extreme outliers may be missed. Therefore, to the extent that such extreme outliers are important for benchmarking studies, these would have to be inserted into the datasets that we have generated.

[1] The OSF repository is available here: https://doi.org/10.17605/OSF.IO/SMJTZ

Our generated datasets are limited to 256 columns. This is reflective of most studies that utilize clinical and administrative health data. However, for benchmarks requiring very high dimensional datasets, the current generative model would not be able to accommodate that.

## Acknowledgements

The authors declare the use of generative AI in the research and writing process. According to the GAIDeT taxonomy (2025), the following tasks were delegated to GAI tools under full human supervision:

- Proofreading and editing

The GAI tool used was: ChatGPT 5. Responsibility for the final manuscript lies entirely with the authors. GAI tools are not listed as authors and do not bear responsibility for the final outcomes.

## Author contributions

Study design: HY, FD, KEE. Data preparation and curation: HY, LP, DL. Development and analysis: HY, FD, KEE. Writing and reviewing manuscript: HY, FD, KEE, LK, LP, DL. Funding and supervision: KEE, LK.

## Funding statement

This research is funded by the Canada Research Chairs program through the Canadian Institutes of Health Research, a Discovery Grant RGPIN-2022-04811 from the Natural Sciences and Engineering Research Council of Canada, a CIFAR catalyst grant, and by the Canadian Children Inflammatory Bowel Disease Network. LP was funded by the Deutsche Forschungsgemeinschaft (DFG, German Research Foundation) – 530282197.

## Competing Interests Statement

KEE and LP have financial interests in Woodway Assurance, a spin-off company from their academic research lab that develops privacy-enhancing technologies.

# Appendix

## A1. Datasets

In our experiments, we use the public OpenML datasets and 13 healthcare related datasets. The additional 13 datasets contain medical records related to inpatient, discharge, pregnancy, COVID-19, and health surveys, with details listed in Table 3.

**Table 3:** Details of the 13 healthcare related datasets.

| **Datasets** | **# Columns** | **# Records** |
|---|---|---|
| Better Outcomes Registry & Network (BORN) | 121 | 968,435 |
| Basic Stand Alone (BSA) Inpatient Claims | 7 | 588,415 |
| California Hospital Discharge (California 2007) | 402 | 4,017,998 |
| Canadian Community Health Survey (CCHS) | 134 | 904,813 |
| Canadian COVID-19 (COVID) | 11 | 1,284,881 |
| FDA Adverse Event Reporting System (FAERS) | 36 | 881,204 |
| Florida Hospital Discharge (Florida 2007) | 303 | 2,563,370 |
| Medical Information Mart for Intensive Care III (MIMIC-III) | 17 | 30,662 |
| New York Hospital Discharge (New York 2007) | 330 | 2,608,615 |
| COVID Survival (Nexoid) | 55 | 968,408 |
| Texas Inpatient Data (Texas) | 75 | 745,999 |
| Washington State Hospital Discharge (Washington 2007) | 357 | 644,902 |
| Washington State Hospital Discharge (Washington2008) | 424 | 652,344 |